\documentclass[conference]{IEEEtran}

\usepackage{cite}
\usepackage{hyperref}
\usepackage{balance}
\usepackage{color}
\usepackage[none]{hyphenat}

\begin{document}
	


\title{MediaEval 2018: Predicting Media Memorability}

\author{\IEEEauthorblockN{Romain Cohendet\IEEEauthorrefmark{1},
		Claire-H\'el\`ene Demarty\IEEEauthorrefmark{1},
		Ngoc Q. K. Duong\IEEEauthorrefmark{1},
		Mats Sj\"oberg\IEEEauthorrefmark{2},
		Bogdan Ionescu\IEEEauthorrefmark{3} and
		Thanh-Toan Do\IEEEauthorrefmark{4}}
	
		\IEEEauthorblockA{\IEEEauthorrefmark{1} Technicolor, Rennes, France}
		\IEEEauthorblockA{\IEEEauthorrefmark{2} Aalto University, Finland}
		\IEEEauthorblockA{\IEEEauthorrefmark{3} Multimedia Lab, University Politehnica of Bucharest, Romania}
		\IEEEauthorblockA{\IEEEauthorrefmark{4} ARC Center of Excellence for Robotic Vision, University of Adelaide, Australia}}

\maketitle

\begin{abstract}
In this paper, we present the Predicting Media Memorability task, which is proposed as part of the MediaEval 2018 Benchmarking Initiative for Multimedia Evaluation.
Participants are expected to design systems that automatically predict memorability scores for videos, which reflect the probability of a video being remembered.
In contrast to previous work in image memorability prediction, where memorability was measured a few minutes after memorization, the proposed dataset comes with ‘short-term’ and ‘long-term’ memorability annotations.
All task characteristics are described, namely: the task's challenges and breakthrough, the released data set and ground truth, the required participant runs and the evaluation metrics.
\end{abstract}

\vspace{-0.3cm}
\section{Introduction}
Following the rapid expansion of the image memorability prediction's research field \cite{isola_2011_makes,khosla_2015_understanding,baveye_2016_deep,fajtl_2018_amnet}, the challenge has recently been extended to videos \cite{han_2015_learning,shekhar_2017_show,cohendet_2018_annotating}.
An important motivation for video memorability (VM) prediction derives from the need for new techniques that can help to organize and retrieve digital content, to make it more useful in our daily lives.
The problem is a pressing one since media platforms, such as social networks, search engines, and recommender systems deal with growing amounts of content data day after day. 
Like other cues of video importance, such as aesthetics or interestingness, memorability can be regarded as useful to help make a choice between otherwise comparable videos. 
Consequently, a large number of applications, e.g., education and learning, content retrieval and search, content summarization, storytelling, targeted advertising, content recommendation and filtering, would benefit from models capable of ranking videos according to their memorability.

Despite its potential of being an active area of reseach in the computer vision community, VM prediction suffers from two main obstacles that were described in \cite{cohendet_2018_annotating}. 
Firstly, among the previous attempts at predicting VM \cite{han_2015_learning,shekhar_2017_show,cohendet_2018_annotating} no clear definition of VM arises, nor does a common and unified protocol for its measurement exist, contrary to what can be found in the literature for image memorability. Secondly, no large-scale dataset is available, for the community to build its models.
The purpose of this task is therefore to propose a public benchmark to assess the memorability of videos, based on a publicly released large-scale dataset and on an objective and clear measurement protocol.

\vspace{-0.3cm}
\section{Task description}
The Predicting Media Memorability Task requires participants to build systems that are capable of predicting how memorable a video is, by computing for each video a memorability score. Participants will be provided with an extensive dataset of videos with memorability annotations.
The ground truth has been collected through recognition tests, and, for this reason, reflects objective measures of memory performance.
In contrast to previous work on image memorability prediction \cite{isola_2011_makes,khosla_2015_understanding}, where memorability was measured a few minutes after memorization, the dataset comes with both ‘short-term’ and ‘long-term’ memorability annotations. 
Because memories continue to evolve in long-term memory \cite{mcgaugh_2000_memory}, in particular during the first day following memorization (see e.g., the forgetting curve in the seminal work of Ebbinghaus \cite{murre_2015_replication}), we expect long-term memorability annotations to be more representative of long-term memory performance, which is more relevant in many applications.

Participants will be required to train computational models capable of inferring video memorability from visual content.
Optionally, descriptive titles attached to the videos may be used. Two subtasks will be offered to participants:
\begin{itemize}
	\item \textit{\bf Short-term Memorability Subtask}: the task involves predicting a ‘short-term’ memorability score for a given video.
	\item \textit{\bf Long-term Memorability Subtask}: the task involves predicting a ‘long-term’ memorability score for a given video.
\end{itemize}
For the two subtasks, depending on the runs, participants will be allowed to use external data.

\vspace{-0.3cm}
\section{Data description}
The dataset is composed of 10,000 short soundless videos shared under a license that allows their use and redistribution in the context of MediaEval 2018.
These 10,000 videos were split into 8,000 videos for the development set and 2,000 videos for the test set.
They were extracted from raw footage used by professionals when creating content. Of 7s-duration each, they are varied and contain different scenes types.
Each video also comes with its original title.
These titles can often be seen as a list of tags (textual metadata) that might be useful to infer the memorability of the videos.
Participants are free to use them or not.

To facilitate participation from various communities, we also provide some pre-computed content descriptors.
Two of them are video-dedicated features: \textit{C3D spatio-temporal visual features} that are obtained by extracting the output of the final classification layer of the C3D model, a 3-dimensional convolutional network proposed for generic video analysis \cite{tran_2015_c3d}, and \textit{HMP} \cite{almeida_2011_comparison}, the histogram of motion patterns for each video. Additional frame-based features are provided that were extracted on three key-frames (first, middle and last frames) for each video: \textit{HoG descriptors} (Histograms of Oriented Gradients) \cite{dalal_2005_histograms} are calculated on 32x32 windows on a grey scale version of each frame; \textit{LBP} (Local Binary Patterns) \cite{he_1990_texture} are calculated for patches of 8x15 pixels; \textit{InceptionV3 features} \cite{szegedy_2016_rethinking} correspond to the output of the fc7 layer of the InceptionV3 deep network; \textit{ORB features} \cite{rublee_2011_orb} result from a fusion of FAST keypoint detector and BRIEF descriptor and \textit{Color histograms} are computed in the HSV space. Additionally, following the work in \cite{haas_2015_can}, a set of \textit{Aesthetic visual features}, composed of color, texture and object based descriptors, aggregated through the computation of their mean and median values, are extracted for each 10-frame of one single video.

\vspace{-0.3cm}
\section{Ground truth}
\subsection{Protocol to measure video memorability}
To collect memorability annotations, we introduced a new protocol to measure human long-term memory performance for videos, partly inspired by \cite{isola_2011_makes}.
The protocol consists of two parts, and is based upon recognition tests for our memorability scores to reflect objective measures of memory performance.
The first part consists of interlaced viewing and recognition tasks.
Participants viewed a sequence of 180 videos, 40 of which being \textit{targets} videos, i.e., repeated videos, and the other being fillers, i.e., videos that occurred only once.
Their task was to press the space bar whenever they detected a repetition.
After 24 to 72 hours, participants viewed a new sequence of videos consisting of 40 targets, which were videos randomly chosen from the fillers of the first part, and 120 new fillers.
In contrast to previous work on image memorability prediction \cite{isola_2011_makes,khosla_2015_understanding}, where memorability was measured a few minutes after memorization, memory performance was therefore measured twice: a few minutes after memorization and again (on different items) 24-72 hours later.
Thus, the dataset comes with both short-term and long-term memorability annotations.
These two scores will allow a comparison of the participants' systems for both short and long term memorability prediction.
However, because of the difficulty to collect data after a long delay through crowdsourcing, the number of annotations is bigger for short-term memorability scores than for long-term ones.
On average, each video received 38 annotations in the short-term recognition task and 13 annotations in the long-term recognition task.
For each video in the development set, we provide the number of annotations for both the short-term and long-term recognition tasks.

\subsection{Memorability scores calculation}
We assigned an initial memorability score to each video, defined as the percentage of correct detections by participants, for both short-term and long-term memory performances.

The short-term raw scores are further refined by applying a linear transformation that takes into account the memory retention duration to correct/normalize the scores.
Indeed, in our measurement protocol, the second occurrence (i.e., repetition) of a video happens after variable time intervals (i.e., each video is repeated after a variable number of other videos randomly chosen in the range of [45;100] videos).
In \cite{isola_2014_makes}, using a similar approach for images, it has been shown that memorability scores evolve as a function of the time interval between repeats while memorability ranks are largely conserved. We were able to prove the same relation for videos.
Thus, as in \cite{khosla_2015_understanding}, we use this information to apply a correction to our raw memorability scores to explicitly account for the difference in interval lengths, with the objective for our short-term memorability scores to be the most representative of the typical memory performance after the max interval (i.e., 100 videos).
Because we observed that memorability decreases linearly when the retention duration increases, we decided to apply a linear correction.
Nevertheless, note that the applied correction only has a little effect on the memorability scores both in term of absolute and relative values.

On the contrary, we did not apply any correction for long-term memorability scores.
Indeed, we observed no specific relationship between retention duration and long-term memorability from our collected scores.
This was expected from what can be found in the literature: according to our protocol, the second measure was carried out 24 to 72 hours after the first measure.
After such a long retention duration, it is expected that the memory performance is no more subjected to substantial decrease due to the retention duration.

\vspace{-0.3cm}
\section{Run description}
Every team can submit up to 10 runs, 5 per subtask.
For each subtask, a required run is defined. \textit{\bf Short-term memorability subtask -- required run}: Any information (extracted from the content, the provided features, the short-term memorability scores or external data) is allowed to build the systems, except the use of the long-term memorability scores which is not allowed.
\textit{\bf Long-term memorability subtask -- required run}: Any information (extracted from the content, the provided features, the long-term memorability scores or external data) is allowed to build the systems, except the use of the short-term memorability scores which is not allowed.
Apart from these required runs, any additional run for each subtask will be considered as a general run, i.e., anything
is allowed, both from the method point of view and the information sources.

\vspace{-0.3cm}
\section{Evaluation}
For both subtasks, the official evaluation metric will be the Spearman's rank correlation between the predicted memorability scores and the ground-truth memorability scores computed over all test videos.
Although the task remains a prediction task, only the ranking of the different videos will be evaluated by the official metric.
The choice of the Spearman's rank correlation as official measure indeed corresponds to a desire of normalizing the output of the different systems and making the comparison easier.
For this reason, participants are encouraged to really consider the task as a prediction task.
Other classic metrics will also be computed and provided to the participants for the sake of comparison between the different runs and systems.

\vspace{-0.3cm}
\section{Conclusions}
A complete and comparative framework for the evaluation of video memorability is proposed.
Details on the methods and results of each individual participant team can be found in the working note papers of the MediaEval 2018 workshop proceedings.

\vspace{-0.3cm}
\section{Acknowledgments}
\small{We would like to thank Ricardo Manhaes Savii (Federal University of São Paulo) and Mihai Gabriel Constantin (University Politehnica of Bucharest) for providing the features that accompany the released data.}
\small{Bogdan Ionescu's work was partially supported by the Romanian Ministry of Innovation and Research, UEFISCDI, project SPIA-VA, agreement 2SOL/2017, grant PN-III-P2-2.1-SOL-2016-02-0002}
\small{Mats Sj\"{o}berg's work has been supported by the Academy of Finland in the project 313988 and the European Union's Horizon 2020 Research and Innovation Programme under Grant Agreement No 780069.}

\bibliographystyle{IEEEtran}
\bibliography{references} 
\end{document}